\documentclass[11pt]{article}

\usepackage{acl}

\usepackage{times}
\usepackage{latexsym}
\usepackage[T1]{fontenc}
\usepackage[utf8]{inputenc}
\usepackage{microtype}
\usepackage{inconsolata}

\usepackage{graphicx}
\usepackage{subfigure}
\usepackage{booktabs}
\usepackage[table]{xcolor}
\usepackage{multirow}
\usepackage{amsmath}
\usepackage{amssymb}
\usepackage{placeins}
\usepackage{makecell}
\usepackage{array}
\usepackage{listings}

\definecolor{LightGray}{gray}{0.9}
\definecolor{mycolor-1}{RGB}{220, 242, 247}
\definecolor{mycolor_lightblue}{RGB}{231, 247, 253}
\definecolor{mycolor_lightgreen}{RGB}{239, 251, 240}
\definecolor{mycolor_lightred}{RGB}{253, 246, 241}
\definecolor{mycolor_lightpink}{RGB}{253, 245, 252}
\definecolor{lblue}{RGB}{0, 102, 204}  

\usepackage{hyperref}
\hypersetup{
    colorlinks=true,
    linkcolor=lblue,
    filecolor=magenta,      
    urlcolor=lblue,
    citecolor=lblue
}

\title{HeLa-Mem: Hebbian Learning and Associative Memory for LLM Agents}

\author{
  Jinchang Zhu$^{1,*,a}$, Jindong Li$^{1,*}$, Cheng Zhang$^{2,*}$, Jiahong Liu$^{3}$, Menglin Yang$^{1,\dagger,b}$ \\
  $^{1}$The Hong Kong University of Science and Technology (Guangzhou) \\
  $^{2}$Jilin University \\
  $^{3}$The Chinese University of Hong Kong \\
  \texttt{$^{a}$jzhu997@connect.hkust-gz.edu.cn \quad $^{b}$menglinyang@hkust-gz.edu.cn} \\
  $^{*}$Equal contribution. $^{\dagger}$Corresponding author.
}

\begin{document}

\maketitle

\begin{abstract}
Long-term memory is a critical challenge for Large Language Model agents, as fixed context windows cannot preserve coherence across extended interactions. Existing memory systems represent conversation history as unstructured embedding vectors, retrieving information through semantic similarity. This paradigm fails to capture the \textit{associative structure} of human memory, wherein related experiences progressively strengthen interconnections through repeated co-activation. Inspired by cognitive neuroscience, we identify three mechanisms central to biological memory: \textit{association}, \textit{consolidation}, and \textit{spreading activation}, which remain largely absent in current research.
To bridge this gap, we propose \textbf{HeLa-Mem}, a bio-inspired memory architecture that models memory as a dynamic graph with Hebbian learning dynamics. HeLa-Mem employs a dual-level organization: (1) an \textit{episodic memory graph} that evolves through co-activation patterns, and (2) a \textit{semantic memory store} populated via Hebbian Distillation, wherein a Reflective Agent identifies densely connected memory hubs and distills them into structured, reusable semantic knowledge. This dual-path design leverages both semantic similarity and learned associations, mirroring the episodic-semantic distinction in human cognition. Experiments on LoCoMo demonstrate superior performance across four question categories while using significantly fewer context tokens (Figure~\ref{fig:tradeoff}). Code is available on \href{https://github.com/ReinerBRO/HeLa-Mem}{GitHub}.



\end{abstract}

\section{Introduction}


 Large language models have demonstrated remarkable capabilities in language understanding and generation, enabling increasingly sophisticated interactive agents~\citep{2025_arXiv_Qwen2.5_Qwen2.5-Technical-Report, liuperfit}. However, sustaining coherent behavior over long time horizons remains a fundamental challenge~\citep{2025_TOIS_Survey_A-Survey-on-the-Memory-Mechanism-of-Large-Language-Model-based-Agents}. Due to their reliance on fixed-length context windows, LLMs struggle to maintain consistent representations of past interactions as dialogues extend or span multiple sessions. This limitation often leads to fragmented memory~\citep{2025_arXiv_Survey_From-Human-Memory-to-AI-Memory=A-Survey-on-Memory-Mechanisms-in-the-Era-of-LLMs, 2025_arXiv_Survey_Memory-in-the-Age-of-AI-Agents, 2025_arXiv_Survey_AI-Meets-Brain=Memory-Systems-from-Cognitive-Neuroscience-to-Autonomous-Agents, liu2025personalizedsurvey}, resulting in factual inconsistencies, diminished personalization, and unstable agent behavior~\citep{li2026discrete}. Addressing long-term memory coherence is therefore essential for LLM agents operating in settings that require persistent user adaptation, multi-session knowledge retention, or stable persona maintenance.

\begin{figure}[t]
    \centering
    \includegraphics[width=0.95\columnwidth]{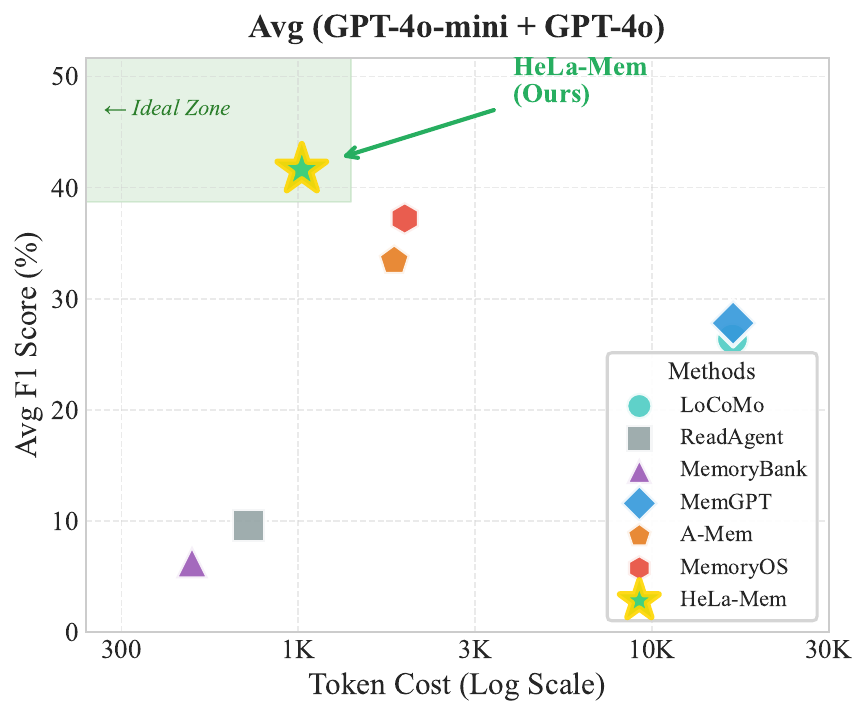}
    \caption{Performance vs. token efficiency on LoCoMo, averaged across GPT-4o-mini and GPT-4o. HeLa-Mem achieves strong performance with fewer tokens, landing in the upper-left ideal region.}
    \label{fig:tradeoff}
\end{figure}

Current memory mechanisms for LLM agents can be broadly categorized into three methodological paradigms.  \textit{Knowledge-organization methods}, such as A-Mem \citep{xu2025mem}, structure memory into interconnected semantic networks to enable adaptive management. \textit{Retrieval mechanism-oriented approaches}, exemplified by MemoryBank \citep{zhong2024memorybank}, integrate semantic retrieval with memory forgetting curves for long-term updating. \textit{Architecture-driven methods}, including MemGPT \citep{packer2023memgpt}, employ hierarchical memory structures with explicit read and write operations to dynamically manage limited context windows. Despite their demonstrated effectiveness, these approaches are typically developed in isolation, each prioritizing a single dimension—memory structure, retrieval strategy, or update mechanism—while largely overlooking their mutual interaction and joint contribution to long-term coherence.

More fundamentally, this component-wise optimization often overlooks a critical aspect of long-term coherence: the \textit{dynamic evolution} of memory structure. Human memory is not a static database where items are stored and retrieved in isolation; rather, it is a dynamic system where connections are continuously reorganized by experience. For example, a topic discussed today might trigger a memory from a month ago, not necessarily because they share surface-level keywords, but because they are part of the same evolving narrative arc. Current systems, by treating storage and retrieval as separate static processes, fail to capture this evolving connectivity, leading to agents that ``remember'' facts but lack the ``continuity'' of a developing relationship.

\begin{figure}[t]
    \centering
    \includegraphics[width=1.0\linewidth]{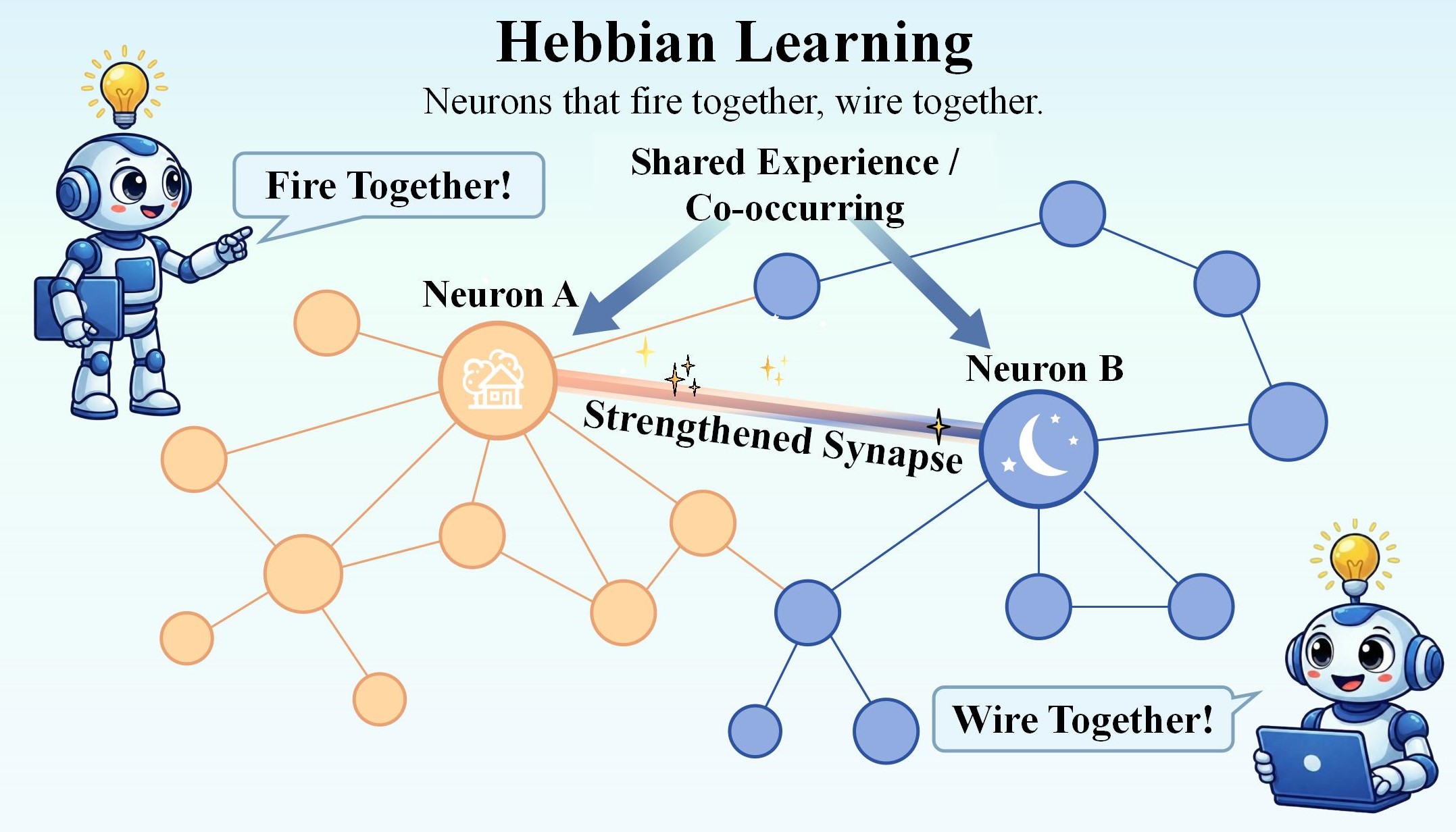}
    \caption{
    Conceptual illustration of Hebbian learning in associative memory. Two memory nodes (Neuron A and B) representing distinct experiences (e.g., a daytime event and a nighttime event) develop strengthened synaptic connections when co-activated through shared context. This "neurons that fire together, wire together" principle forms the theoretical foundation of HeLa-Mem's dynamic memory graph.
    }
    \label{fig:hebbian_concept}
\end{figure}

To better understand how long-term coherence can be maintained, we draw on a fundamental principle of biological memory: \textit{Hebbian learning}. In biological systems, experiences that are repeatedly co-activated gradually develop stronger associations, a phenomenon often summarized as “neurons that fire together wire together” (illustrated in Figure \ref{fig:hebbian_concept}). This associative organization allows related memories to be efficiently reactivated through spreading activation and supports the gradual consolidation of episodic experiences into more stable semantic knowledge. Together, association, consolidation, and spreading activation form a tightly coupled memory process that enables biological systems to maintain coherent representations over extended time scales—capabilities that remain largely absent from current artificial memory designs.

Building on this perspective, we propose HeLa-Mem (\underline{\textbf{He}}bbian \underline{\textbf{L}}earning \underline{\textbf{a}}ssociative \underline{\textbf{Mem}}ory), a unified memory architecture for LLM agents. HeLa-Mem represents conversational history as a dynamic graph with Hebbian learning dynamics and operates through coordinated mechanisms of online association, reflective consolidation, and dual-path retrieval, establishing a unified memory management framework that captures both fine-grained details and high-level patterns.

The primary contributions of our work are:
\begin{itemize}
\item We propose \textbf{HeLa-Mem}, a bio-inspired framework that utilizes an \textit{Online Encoding \& Association} mechanism to model conversation history as a dynamic Hebbian graph, where co-activated memories strengthen connections to capture latent context.
\item We introduce a \textit{Reflective Consolidation} framework using \textbf{Hebbian Distillation}, which identifies hub clusters and transforms them into structured semantic knowledge, preventing graph explosion while retaining key information.
\item We implement a \textit{Dual-Path Retrieval} strategy that leverages spreading activation to traverse Hebbian edges, achieving the best average rank (1.25) across all question categories.
\item Comprehensive experiments on the LoCoMo benchmark validate HeLa-Mem's effectiveness, achieving superior performance across four categories while using significantly fewer context tokens.
\end{itemize}

\section{Related Work}

\subsection{Memory for LLM Agents}

\begin{figure*}[t]
    \centering
    \includegraphics[width=0.99\linewidth]{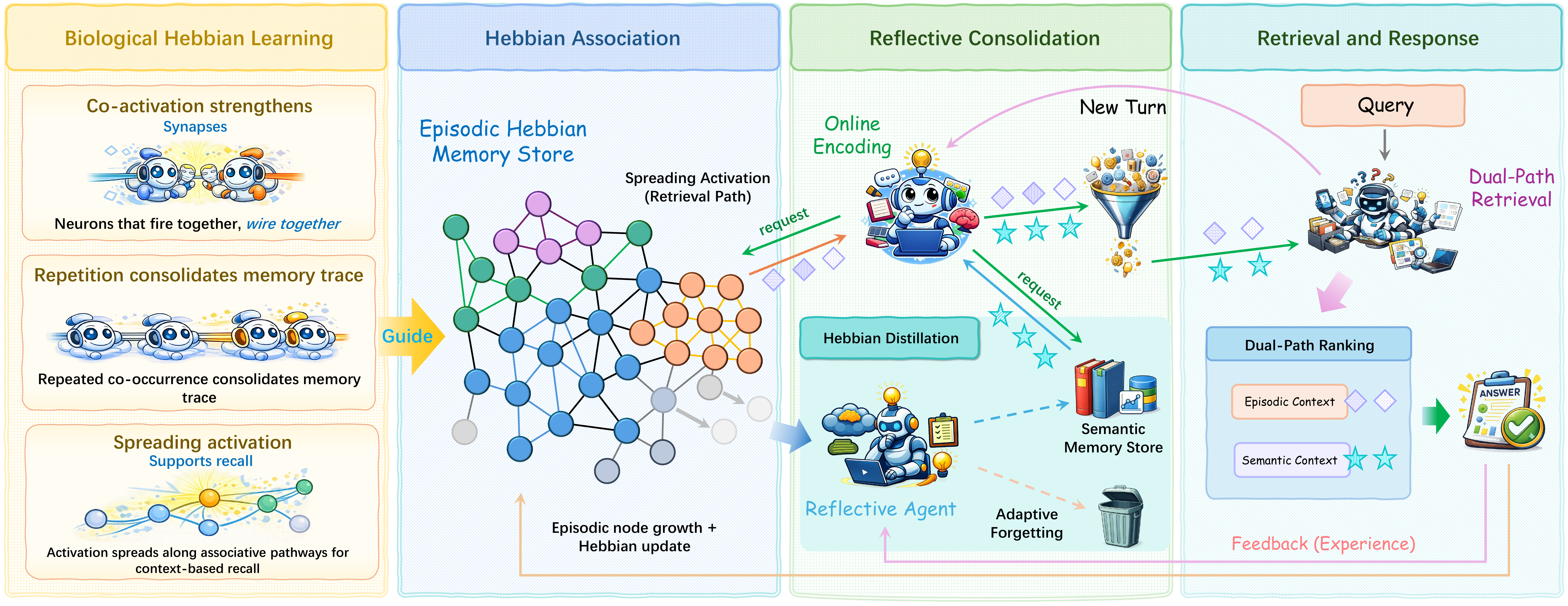}
    \caption{\textbf{The architectural overview of HeLa-Mem.} The framework consists of three modules: (1) \textbf{Hebbian Association} for dynamic graph construction (Section \ref{sec:encoding}); (2) \textbf{Reflective Consolidation} for semantic knowledge distillation (Section \ref{sec:consolidation}); and (3) \textbf{Retrieval and Response} using a Dual-Path strategy (Section \ref{sec:retrieval}).}
    \label{fig:framework}
\end{figure*}

Existing Large Language Models face fundamental challenges in handling complex scenarios requiring long-term coherence \citep{2025_arXiv_Survey_Memory-in-the-Age-of-AI-Agents}. Advancements in memory systems addressing this problem can be broadly grouped into three categories.

Knowledge-organization methods focus on capturing and structuring intermediate reasoning states. Think-in-Memory \citep{liu2023think} stores evolving chains-of-thought, enabling consistency through continual updates. A-Mem \citep{xu2025mem} organizes knowledge into an interconnected note network that spans sessions.

Retrieval mechanism-oriented approaches, pioneered by RAG \citep{lewis2020retrieval}, enrich the model with external memory libraries. MemoryBank \citep{zhong2024memorybank} logs conversations, events, and user traits in a vector database and refreshes them using a forgetting-curve schedule. Generative Agents \citep{park2023generative} keep memories in natural language and add a reflection loop for relevance filtering. EmotionalRAG \citep{huang2024emotional} retrieves memory entries by combining semantic similarity with the agent's emotional state.

Architecture-driven designs alter the core control flow to manage context explicitly. MemGPT \citep{packer2023memgpt} adopts an OS-like hierarchy with dedicated read/write calls. SCM \citep{liang2023scm} introduces dual buffers and a memory controller that gates selective recall. Mem0 \citep{chhikara2025mem0} dynamically extracts and consolidates salient information for scalable long-term memory. MemoryOS \citep{kang2025memory} introduces a three-tier hierarchical storage with short-term, mid-term, and long-term memory units, employing segment-page organization for dynamic updating.

\subsection{Hebbian Learning in Neural Networks}

Hebbian learning, summarized as ``neurons that fire together wire together'' \citep{hebb2005organization}, is a foundational principle in neuroscience describing how synaptic connections strengthen through correlated activity (see Figure \ref{fig:hebbian_concept}). Formally, given activation states $x_i$ and $x_j$ of two neurons, the connection weight $w_{ij}$ is updated as $\Delta w_{ij} = \eta \cdot x_i \cdot x_j$, where $\eta$ is the learning rate. This principle has been applied in Hopfield networks \citep{1982_PNAS_Neural-Networks-and-Physical-Systems-with-Emergent-Collective-Computational-Abilities}, which demonstrate how recurrent neural networks with symmetric connections can function as associative memories, storing and retrieving patterns through energy minimization. More recently, \citet{ramsauer2020hopfield} show that modern Hopfield networks with continuous states are mathematically equivalent to the attention mechanism in Transformers, revealing deep connections between biological memory principles and contemporary deep learning architectures. In the context of LLM agents, Hebbian dynamics offer a principled approach to capture latent associations between memories that may not be apparent from semantic similarity alone.

\section{HeLa-Mem Architecture}

Inspired by the synaptic plasticity of the human brain, HeLa-Mem models conversation history as a dynamic associative graph rather than a static log. Our design is guided by three neuroscience intuitions: (1) \textit{Association over Isolation}---memories that co-occur should wire together, forming latent pathways beyond simple semantic similarity; (2) \textit{Active Consolidation}---frequently accessed memory clusters should solidify into stable knowledge, similar to sleep-based consolidation; and (3) \textit{Spreading Retrieval}---recalling one memory should naturally trigger related concepts through established synaptic routes.

Based on these principles (Figure \ref{fig:framework}), HeLa-Mem operates through a continuous cognitive lifecycle:
\begin{itemize}
    \item \textbf{Online Encoding \& Association}: Conversation turns are encoded into the Episodic Memory Graph, where a Hebbian Learning mechanism dynamically strengthens connections between co-activated memories.
    \item \textbf{Reflective Memory Agent}: Upon reaching associative thresholds, this agent identifies hub nodes and applies \textit{Hebbian Distillation} to consolidate them into stable semantic knowledge, preventing noise accumulation.
    \item \textbf{Dual-Path Retrieval}: During query time, queries activate both specific episodic details and broader semantic knowledge through spreading activation.
\end{itemize}


\subsection{Memory Storage}

\subsubsection{Episodic Memory Graph}

Conversation turns are stored as nodes in a weighted graph. Each node contains the original text, a dense embedding, the timestamp, extracted keywords, and the speaker role. Edges between nodes represent associative connections, with weights indicating the strength of association. Initially, consecutive turns are connected with small weights; these weights evolve through Hebbian learning.

\subsubsection{Semantic Memory Store}

The semantic level stores distilled knowledge extracted from episodic memories:
specifically, we apply \textit{Hebbian Distillation} on hub-centered clusters in the episodic graph to produce \textit{distilled semantic records} with traceable evidence links to their source turns.

\begin{itemize}
\item \textbf{User Model}: Stable user characteristics (e.g., ``enjoys outdoor activities'') with confidence scores and supporting evidence.
\item \textbf{Factual Memory}: Extracted facts with absolute timestamps, such as event dates and relationships.
\item \textbf{Agent Knowledge}: The agent's established persona, preferences, and behavioral patterns.
\end{itemize}

This serves as long-term memory that persists beyond conversation windows.


\subsection{Online Encoding \& Association}
\label{sec:encoding}
Synaptic efficacy in the brain is not fixed; it is plastic, evolving based on activity patterns. We emulate this dynamic through Hebbian learning to capture latent associations that semantic embeddings alone might miss.

Following the neuroscience principle that ``neurons that fire together wire together,'' edge weights strengthen when memories are co-activated during retrieval:
\begin{equation}
w_{ij}^{(t+1)} = \underbrace{(1 - \lambda) \cdot w_{ij}^{(t)}}_{\text{synaptic decay}} + \underbrace{\eta \cdot \mathbb{I}(v_i, v_j \in \mathcal{K}_t)}_{\text{active reinforcement}} ,
\end{equation}
where $\lambda$ is the decay rate, $\eta$ is the learning rate, and $\mathbb{I}(\cdot)$ is an indicator that the pair $(v_i, v_j)$ is co-activated in the current retrieval set $\mathcal{K}_t$. This dynamic allows frequently correlated memories to strengthen while unused connections fade over time.


\subsection{Reflective Memory Agent}
\label{sec:consolidation}
While associative learning happens during active retrieval, long-term memory maintenance relies on active consolidation. To prevent memory overload and crystallize important information, we introduce a Reflective Agent that mimics the brain's sleep-based consolidation process through \textit{Hebbian Distillation}.

\begin{figure}[t]
    \centering
    \includegraphics[width=0.95\columnwidth]{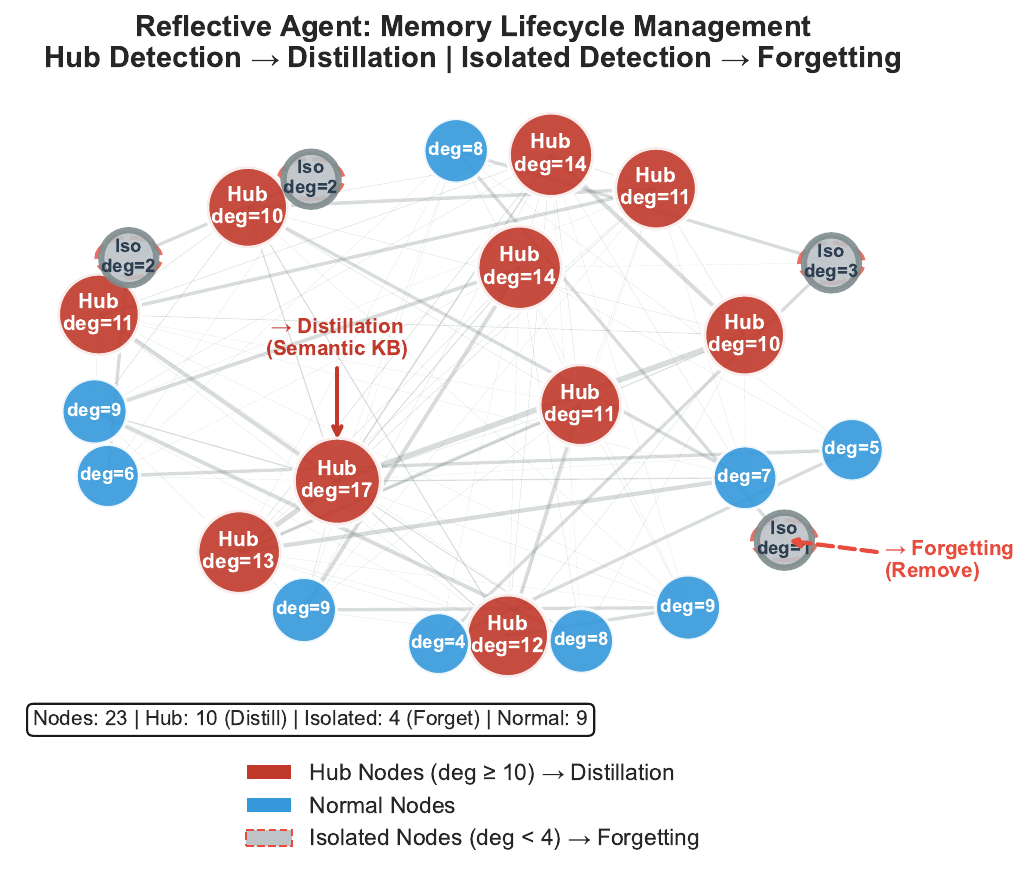}
    \caption{Hebbian memory graph showing the Reflective Agent's dual role. \textbf{Hub nodes} (red, high degree) are candidates for Hebbian Distillation, consolidating related memories into structured semantic knowledge. \textbf{Isolated nodes} (gray, low degree with dashed circles) are candidates for Adaptive Forgetting, maintaining memory efficiency.}
    \label{fig:reflective_dual}
\end{figure}

The Reflective Agent monitors the graph's structural evolution to manage the memory lifecycle (see Figure \ref{fig:reflective_dual}), analogous to sleep-based memory consolidation in the brain.

\textbf{Hub Detection.} Nodes that have accumulated high total edge weight through Hebbian learning are identified as hubs. Specifically, a node $v_i$ is flagged for consolidation if its \textit{associative strength} exceeds a threshold $\delta_{hub}$:
\begin{equation}
D(v_i) = \sum_{j \in \mathcal{N}(i)} w_{ij} > \delta_{hub} .
\end{equation}

Upon detecting that a node's accumulative weight exceeds the threshold $\delta_{hub}$, the agent triggers Hebbian Distillation. To capture the full context, the agent retrieves the hub node along with its strongly connected neighbors. The LLM synthesizes this cluster of related memories to identify common themes and causal relationships, abstracting them into declarative semantic entries. These distilled records are stored in the Semantic Memory Store, effectively compressing repetitive episodic details into stable, generalizable knowledge.

\textbf{Adaptive Forgetting.} This process is triggered when a node's status falls below critical retention thresholds. A memory is flagged for removal only if it simultaneously satisfies three criteria: (1) its total edge weight is below $\delta_{prune}$ (indicating structural irrelevance), (2) its inactive duration exceeds $\delta_{age}$ (indicating temporal dormancy), and (3) it has zero recent access. This strict compound criterion ensures that the system selectively removes noise while preserving strong, albeit older, associations.


\subsection{Dual-Path Retrieval}
\label{sec:retrieval}
Human memory retrieval is rarely a single-step lookup; it is a spreading activation process where one thought triggers another. HeLa-Mem adopts a dual-path retrieval strategy to emulate this interaction between direct recall and associative spreading.

Given a query, retrieval proceeds in two stages.

\textbf{Base Activation.} Each episodic node receives an initial score combining embedding similarity, temporal decay, and keyword overlap:
\begin{equation}
\begin{aligned}
S_{base}(v_i)
  &= \left(\text{sim}(\mathbf{q}, \mathbf{e}_i)
     + \alpha \cdot \text{keyword\_match}\right) \\
  &\quad \cdot \gamma(v_i) ,
\end{aligned}
\end{equation}
where $\text{sim}(\cdot, \cdot)$ denotes cosine similarity between the query embedding $\mathbf{q}$ and node embedding $\mathbf{e}_i$, $\gamma(v_i) = \exp(-\Delta t / \tau)$ is the temporal decay factor with time constant $\tau$, and $\alpha$ controls the bonus for keyword matches.

\textbf{Spreading Activation.} High-scoring nodes propagate activation through Hebbian edges:
\begin{equation}
S(v_j) = S_{base}(v_j) + \beta \sum_{i \in \mathcal{N}(j)} S_{base}(v_i) \cdot w_{ij} ,
\end{equation}
where $\mathcal{N}(j)$ denotes the neighbors of node $v_j$ in the memory graph, $w_{ij}$ is the Hebbian edge weight, and $\beta$ controls the spreading activation strength. This enables retrieval of memories that are semantically distant from the query but strongly associated with initially activated content---particularly beneficial for multi-hop reasoning.

\textbf{Dual-Path Ranking.} The final retrieval set is constructed by combining two ranked lists:
\begin{equation}
\mathcal{R}_{final} = \underbrace{\text{Top-}k(S_{base})}_{\text{base path}} \cup \underbrace{\text{Top-}m(S \mid v \notin \text{Top-}k)}_{\text{flip path}} ,
\end{equation}
where the base path selects the top-$k$ nodes by $S_{base}$, and the flip path promotes up to $m$ additional nodes that rank highest by spreading-augmented score $S$ but were not already selected. This dual-path approach ensures retrieval of both semantically relevant memories and associatively linked memories that spreading activation surfaces. Semantic memory entries are also retrieved and merged to form the final context.


\subsection{Response Generation}

The LLM generates responses using the integrated context (episodic memories and semantic knowledge) along with a system prompt that establishes the conversational role. The detailed prompt structure is provided in Appendix \ref{app:prompts}.

\section{Experiments}

\subsection{Experimental Settings}

\textbf{Dataset.} We conduct experiments on the LoCoMo benchmark \citep{maharana2024evaluating}, specifically designed for assessing long-term conversational memory capabilities. It consists of ultra-long dialogues averaging 300 turns and about 9K tokens per conversation. Questions span multiple categories to systematically evaluate memory abilities.

\textbf{Evaluation Metrics.} Following prior work on long-term conversational memory \citep{maharana2024evaluating, xu2025mem}, we employ standard F1 and BLEU-1 scores to evaluate performance.

\textbf{Compared Methods.} We compare HeLa-Mem with representative memory methods including LoCoMo (Native), ReadAgent \citep{lee2024human}, MemoryBank \citep{zhong2024memorybank}, MemGPT \citep{packer2023memgpt}, A-Mem \cite{xu2025mem}, Mem0 \citep{chhikara2025mem0}, LightMem \citep{fang2025lightmem}, and MemoryOS \citep{kang2025memory}. Baseline results for GPT-4o-mini, GPT-4o, and Qwen2.5-3b are reported from \citet{xu2025mem}; results for Qwen2.5-14b are reported from \citet{yan2025general}. MemoryOS results marked with $^\dagger$ are reproduced by us.

\textbf{Implementation Details.} We evaluate HeLa-Mem across four backbone LLMs: GPT-4o-mini, GPT-4o \citep{achiam2023gpt}, Qwen2.5-14b, and Qwen2.5-3b \citep{2025_arXiv_Qwen2.5_Qwen2.5-Technical-Report}. For HeLa-Mem, the time decay constant is set to $\tau = 60$ days. Episodic retrieval uses $k = 10$, and semantic retrieval uses $k = 5$. The Hebbian learning rate is $\eta = 0.02$, edge decay rate $\lambda = 0.995$, spreading activation strength $\beta = 0.1$, and spreading threshold $\theta = 0.6$.

\subsection{Main Results}

Table \ref{tab:main_results} presents the detailed performance breakdown across four different LLM backbones: GPT-4o-mini, GPT-4o, Qwen2.5-14b, and Qwen2.5-3b. Results show that HeLa-Mem consistently outperforms baselines across varying model sizes and capabilities.

\begin{table*}[!t]
\caption{Experimental results on LoCoMo dataset of QA tasks across four categories (Multi Hop, Temporal, Open Domain, and Single Hop) using different methods. Results are reported in F1 and BLEU-1 (\%) scores. Best performance per model is marked in bold. Missing baselines are marked with ``-'' (Token Length (\textcolor{red}{$\downarrow$}): lower values are better; $^\dagger$: Reproduced results).}
\label{tab:main_results}
\centering
\renewcommand{\arraystretch}{1.02}
\scalebox{0.62}{
\begin{tabular}{
  >{\centering\arraybackslash}p{1.8cm}
  >{\centering\arraybackslash}p{3.6cm}
  |>{\centering\arraybackslash}p{1.6cm}
  >{\centering\arraybackslash}p{1.6cm}
  |>{\centering\arraybackslash}p{1.6cm}
  >{\centering\arraybackslash}p{1.6cm}
  |>{\centering\arraybackslash}p{1.6cm}
  >{\centering\arraybackslash}p{1.6cm}
  |>{\centering\arraybackslash}p{1.6cm}
  >{\centering\arraybackslash}p{1.6cm}
  |>{\centering\arraybackslash}p{2.6cm}
}

\toprule
& & \multicolumn{2}{c|}{\textbf{Multi Hop}} & \multicolumn{2}{c|}{\textbf{Temporal}} & \multicolumn{2}{c|}{\textbf{Open Domain}} & \multicolumn{2}{c|}{\textbf{Single Hop}} & \textbf{Token} \\
\textbf{Model} & \textbf{Method} & F1 & BLEU & F1 & BLEU & F1 & BLEU & F1 & BLEU & \textbf{Length (\textcolor{red}{$\downarrow$})} \\

\midrule

\multirow{7}{*}{\rotatebox{90}{GPT-4o-mini}} 
& LoCoMo & 25.02 & 19.75 & 18.41 & 14.77 & 12.04 & 11.16 & 40.36 & 29.05 & 16,910 \\
& ReadAgent & 9.15 & 6.48 & 12.60 & 8.87 & 5.31 & 5.12 & 9.67 & 7.66 & 643 \\
& MemoryBank & 5.00 & 4.77 & 9.68 & 6.99 & 5.56 & 5.94 & 6.61 & 5.16 & 432 \\
& MemGPT & 26.65 & 17.72 & 25.52 & 19.44 & 9.15 & 7.44 & 41.04 & 34.34 & 16,977 \\
& A-Mem & 27.02 & 20.09 & 45.85 & 36.67 & 12.14 & 12.00 & 44.65 & 37.06 & 2,520 \\
& MemoryOS$^\dagger$ & 38.39 & 29.52 & 41.58 & 35.99 & 23.75 & 17.17 & 45.86 & 40.70 & 2,000 \\

& \cellcolor{mycolor-1}\textbf{HeLa-Mem} & \cellcolor{mycolor-1}\textbf{40.14} & \cellcolor{mycolor-1}\textbf{31.26} & \cellcolor{mycolor-1}\textbf{47.29} & \cellcolor{mycolor-1}\textbf{41.28} & \cellcolor{mycolor-1}\textbf{29.70} & \cellcolor{mycolor-1}\textbf{23.45} & \cellcolor{mycolor-1}\textbf{51.89} & \cellcolor{mycolor-1}\textbf{46.25} & \cellcolor{mycolor-1}\textbf{1,010} \\

\midrule

\multirow{7}{*}{\rotatebox{90}{GPT-4o}} 
& LoCoMo & 28.00 & 18.47 & 9.09 & 5.78 & 16.47 & 14.80 & \textbf{61.56} & \textbf{54.19} & 16,910 \\
& ReadAgent & 14.61 & 9.95 & 4.16 & 3.19 & 8.84 & 8.37 & 12.46 & 10.29 & 805 \\
& MemoryBank & 6.49 & 4.69 & 2.47 & 2.43 & 6.43 & 5.30 & 8.28 & 7.10 & 569 \\
& MemGPT & 30.36 & 22.83 & 17.29 & 13.18 & 12.24 & 11.87 & 60.16 & 53.35 & 16,987 \\
& A-Mem & 32.86 & 23.76 & 39.41 & 31.23 & 17.10 & 15.84 & 48.43 & 42.97 & 1,216 \\
& MemoryOS$^\dagger$ & \textbf{40.23} & \textbf{31.89} & 43.57 & 33.55 & 20.58 & 15.85 & 43.85 & 39.03 & 2,000 \\

& \cellcolor{mycolor-1}\textbf{HeLa-Mem} & \cellcolor{mycolor-1}39.12 & \cellcolor{mycolor-1}29.82 & \cellcolor{mycolor-1}\textbf{50.79} & \cellcolor{mycolor-1}\textbf{44.54} & \cellcolor{mycolor-1}\textbf{24.38} & \cellcolor{mycolor-1}\textbf{19.24} & \cellcolor{mycolor-1}49.69 & \cellcolor{mycolor-1}44.08 & \cellcolor{mycolor-1}\textbf{1,036} \\
\midrule

\multirow{5}{*}{\rotatebox{90}{Qwen2.5-14b}} 
& A-Mem & 22.09 & 15.28 & 27.19 & 22.05 & 13.49 & 10.74 & 33.75 & 30.04 & 1,300 \\
& MEM0 & 31.73 & 24.82 & 28.96 & 26.24 & 15.03 & 11.28 & 42.58 & 35.15 & - \\
& MemoryOS & \textbf{38.19} & \textbf{29.26} & 32.24 & 27.86 & 20.27 & 15.94 & 46.33 & 41.62 & - \\
& LightMem & 25.45 & 19.61 & 32.03 & 27.70 & 15.81 & 11.81 & 34.92 & 31.22 & - \\
& \cellcolor{mycolor-1}\textbf{HeLa-Mem} & \cellcolor{mycolor-1}36.59 & \cellcolor{mycolor-1}27.02 & \cellcolor{mycolor-1}\textbf{36.08} & \cellcolor{mycolor-1}\textbf{29.91} & \cellcolor{mycolor-1}\textbf{24.22} & \cellcolor{mycolor-1}\textbf{20.23} & \cellcolor{mycolor-1}\textbf{49.95} & \cellcolor{mycolor-1}\textbf{45.15} & \cellcolor{mycolor-1}\textbf{944} \\

\midrule

\multirow{7}{*}{\rotatebox{90}{Qwen2.5-3b}} 
& LoCoMo & 4.61 & 4.29 & 3.11 & 2.71 & 4.55 & 5.97 & 7.03 & 5.69 & 16,910 \\
& ReadAgent & 2.47 & 1.78 & 3.01 & 3.01 & 5.57 & 5.22 & 3.25 & 2.51 & 776 \\
& MemoryBank & 3.60 & 3.39 & 1.72 & 1.97 & 6.63 & 6.58 & 4.11 & 3.32 & 298 \\
& MemGPT & 5.07 & 4.31 & 2.94 & 2.95 & 7.04 & 7.10 & 7.26 & 5.52 & 16,961 \\
& A-Mem & 18.23 & 11.94 & 24.32 & 19.74 & \textbf{16.48} & \textbf{14.31} & 23.63 & 19.23 & 1,300 \\
& MemoryOS$^\dagger$ & 19.20 & 14.84 & 20.85 & 16.05 & 13.57 & 10.86 & 25.65 & 18.78 & 2,000 \\

& \cellcolor{mycolor-1}\textbf{HeLa-Mem} & \cellcolor{mycolor-1}\textbf{20.12} & \cellcolor{mycolor-1}\textbf{14.59} & \cellcolor{mycolor-1}\textbf{24.79} & \cellcolor{mycolor-1}\textbf{21.35} & \cellcolor{mycolor-1}12.24 & \cellcolor{mycolor-1}10.24 & \cellcolor{mycolor-1}\textbf{29.51} & \cellcolor{mycolor-1}\textbf{25.91} & \cellcolor{mycolor-1}\textbf{1,072} \\

\bottomrule

\end{tabular}
}
\end{table*}

\textbf{Detailed Analysis (GPT-4o-mini).} Focusing on GPT-4o-mini as a representative case, HeLa-Mem demonstrates significant advantages. In Multi-hop reasoning, it achieves 40.14\%, outperforming MemoryOS (38.39\%) and A-Mem (27.02\%). This validates the Hebbian graph's ability to bridge disparate information pieces through learned associations.

For Temporal tasks, HeLa-Mem scores 47.29\%, surpassing MemoryOS (41.58\%). The preservation of absolute timestamps during distillation allows accurate grounding of relative time expressions. In Open Domain questions, it reaches 29.70\%, providing useful semantic context even for topics outside the main conversation flow. HeLa-Mem leads in Single-hop tasks (51.89\%), demonstrating that the hierarchical retrieval approach remains effective for straightforward factual queries.

\textbf{Token Efficiency.} Notably, HeLa-Mem achieves these results using only $\sim$1,010 tokens on average. This efficiency stems from the selective nature of Hebbian retrieval, which surfaces only the most strongly associated memories without the computational overhead of processing full context windows.

\textbf{Robustness Across Backbones.} To validate stability, Table \ref{tab:avg_results} summarizes the averaged performance across all backbones. HeLa-Mem achieves the best \textbf{Average Rank of 1.25}, significantly surpassing MemoryOS (2.25). This confirms that the theoretical advantages of Hebbian dynamics translate into robust empirical gains regardless of the underlying LLM's scale.

\begin{table}[!t]
\caption{Averaged results across three backbone LLMs (GPT-4o-mini, GPT-4o, Qwen2.5-3b). Avg Rank is computed across all categories lower is better ($^\dagger$Reproduced results).}
\label{tab:avg_results}
\centering
\renewcommand{\arraystretch}{1.3}
\scalebox{0.55}{
\begin{tabular}{l|cc|cc|cc|cc|c}

\toprule

\textbf{Method} & \multicolumn{2}{c|}{\textbf{Multi-hop}} & \multicolumn{2}{c|}{\textbf{Temporal}} & \multicolumn{2}{c|}{\textbf{Open}} & \multicolumn{2}{c|}{\textbf{Single}} & \textbf{Avg} \\
& F1 & BL & F1 & BL & F1 & BL & F1 & BL & \textbf{Rank} \\

\midrule

LoCoMo & 19.21 & 14.17 & 10.20 & 7.75 & 11.02 & 10.64 & 36.32 & 29.64 & 5.00 \\

ReadAgent & 8.74 & 6.07 & 6.59 & 5.02 & 6.57 & 6.24 & 8.46 & 6.82 & 6.38 \\

MemoryBank & 5.03 & 4.28 & 4.62 & 3.80 & 6.21 & 5.94 & 6.33 & 5.19 & 6.88 \\

MemGPT & 20.69 & 14.95 & 15.25 & 11.86 & 9.48 & 8.80 & 36.15 & 31.07 & 4.50 \\

A-Mem & 26.04 & 18.60 & 36.53 & 29.21 & 15.24 & 14.05 & 38.90 & 33.09 & 3.00 \\

MemoryOS$^\dagger$ & 32.61 & 25.42 & 35.33 & 28.53 & 19.30 & 14.63 & 38.45 & 32.84 & 2.25 \\

\cellcolor{mycolor-1}\textbf{HeLa-Mem} & \cellcolor{mycolor-1}\textbf{33.13} & \cellcolor{mycolor-1}\textbf{25.22} & \cellcolor{mycolor-1}\textbf{40.96} & \cellcolor{mycolor-1}\textbf{35.72} & \cellcolor{mycolor-1}\textbf{22.11} & \cellcolor{mycolor-1}\textbf{17.64} & \cellcolor{mycolor-1}\textbf{43.70} & \cellcolor{mycolor-1}\textbf{38.75} & \cellcolor{mycolor-1}\textbf{1.25} \\

\bottomrule

\end{tabular}%
}
\end{table}

\begin{figure*}[t]
    \centering
    \includegraphics[width=1.0\linewidth]{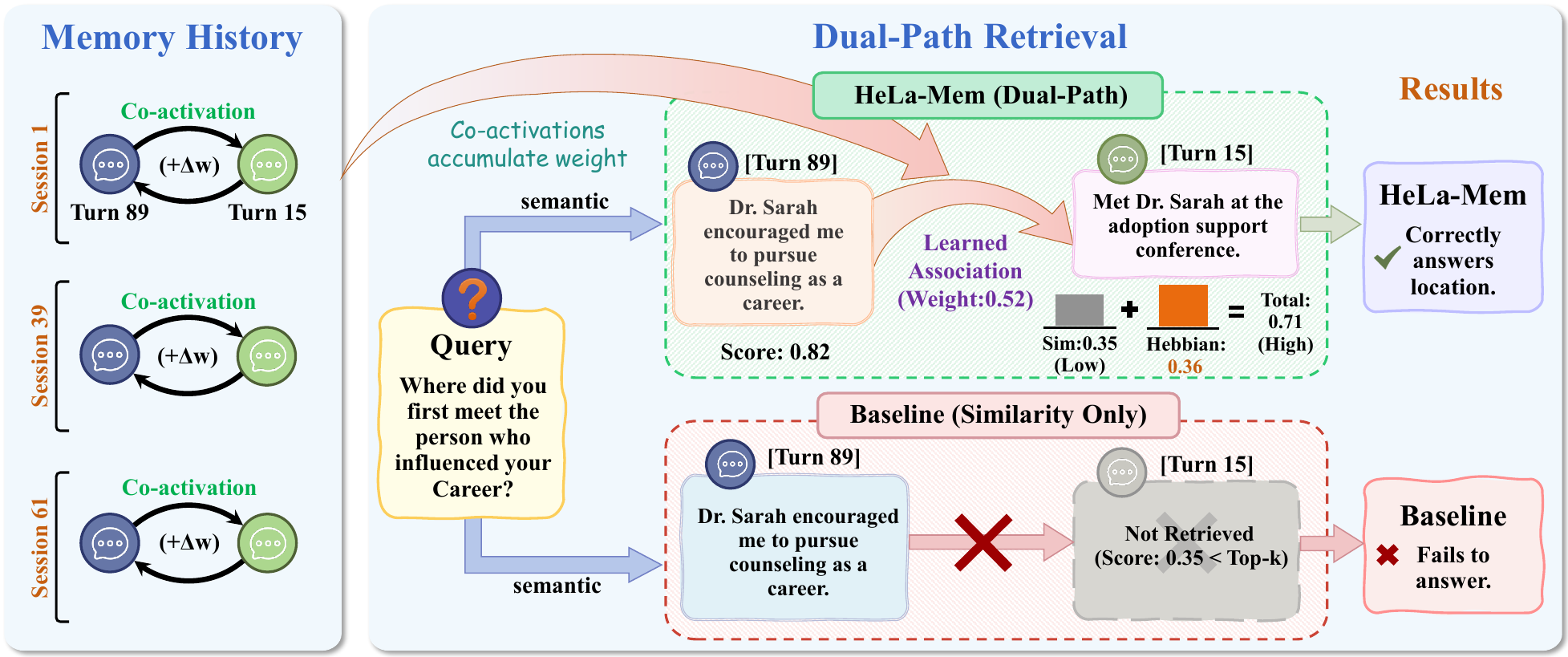}
    \caption{Dual-Path Retrieval for multi-hop reasoning. Given a query requiring both ``career influence'' and ``meeting location,'' the semantic path retrieves Turn 89 (career context). The Hebbian path then propagates activation through learned associations (edge weight 0.52) to retrieve Turn 15 (location context), which semantic similarity alone would miss. The baseline without Spreading Activation cannot bridge these memories.}
    \label{fig:case_study}
\end{figure*}

\subsection{Ablation Study}

To understand the contribution of each component in HeLa-Mem, we conduct ablation experiments by removing key modules. Table \ref{tab:ablation} presents the results on GPT-4o-mini.


\begin{table}[!t]
\caption{Ablation study on LoCoMo benchmark. Results show F1 / BLEU-1 scores (\%).}
\label{tab:ablation}
\centering
\renewcommand{\arraystretch}{1.65}
\scalebox{0.48}{
\begin{tabular}{l|cc|cc|cc|cc|c}

\toprule

\textbf{Variant} & \multicolumn{2}{c|}{\textbf{Multi-hop}} & \multicolumn{2}{c|}{\textbf{Temporal}} & \multicolumn{2}{c|}{\textbf{Open}} & \multicolumn{2}{c|}{\textbf{Single}} & \textbf{\makecell{Avg \\F1}} \\

& F1 & BL & F1 & BL & F1 & BL & F1 & BL & \\

\midrule

\cellcolor{mycolor-1}\textbf{HeLa-Mem (Full)} & \cellcolor{mycolor-1}\textbf{36.04} & \cellcolor{mycolor-1}\textbf{26.56} & \cellcolor{mycolor-1}\textbf{46.23} & \cellcolor{mycolor-1}\textbf{40.48} & \cellcolor{mycolor-1}29.50 & \cellcolor{mycolor-1}23.55 & \cellcolor{mycolor-1}\textbf{45.04} & \cellcolor{mycolor-1}\textbf{39.80} & \cellcolor{mycolor-1}\textbf{34.74} \\

w/o Forgetting & 36.71 & 27.95 & 46.50 & 40.91 & \textbf{30.58} & \textbf{24.45} & 45.24 & 40.01 & 34.28 \\
w/o Spreading Activation & 33.88 & 25.57 & 44.36 & 39.62 & 27.76 & 22.28 & 43.34 & 38.33 & 32.19 \\
w/o Reflective Agent & 30.17 & 22.38 & 42.19 & 36.92 & 24.51 & 19.83 & 40.46 & 34.07 & 29.87 \\

\bottomrule

\end{tabular}%
}
\end{table}

\textbf{Effect of Reflective Agent.} Removing the Reflective Memory Agent causes the largest performance drop (34.74\% $\rightarrow$ 29.87\%), with Multi-hop reasoning suffering most severely (36.04\% $\rightarrow$ 30.17\%). This confirms that the meta-cognitive component is essential for identifying high-degree hub nodes and triggering Hebbian Distillation, which consolidates related episodic memories into structured semantic knowledge.

\textbf{Effect of Spreading Activation.} Disabling spreading activation leads to a notable performance decline (34.74\% $\rightarrow$ 32.19\%), particularly affecting Multi-hop reasoning (36.04\% $\rightarrow$ 33.88\%). This validates our dual-path design: without spreading activation, the system degrades to a single semantic path, failing to retrieve memories that are semantically distant from the query but strongly associated through Hebbian connections, which is crucial for multi-hop reasoning that requires bridging disparate pieces of information. This underscores the value of leveraging historically learned pathways rather than relying solely on static semantic similarity.

\textbf{Effect of Adaptive Forgetting.} Interestingly, removing the forgetting mechanism shows minimal impact on the current LoCoMo benchmark. We attribute this to the limited conversation length ($\sim$300 turns), which does not yet saturate the memory capacity. However, forgetting is critical for scalability in reliable agent deployment. Without it, the memory store would grow unboundedly, inevitably increasing retrieval costs and introducing noise from obsolete information. This mechanism ensures that the system's performance remains stable regardless of conversation duration, acting as a garbage collection process for irrelevant associations.

\subsection{Reflective Agent: Memory Lifecycle Management}

Figure \ref{fig:reflective_dual} illustrates the structure of the Hebbian memory graph after encoding a multi-session conversation. The graph comprises 23 episodic memory nodes, where edge thickness reflects the association strength accumulated through co-activation during retrieval. Node appearance encodes the lifecycle status assigned by the Reflective Agent:

\textbf{Hub Nodes (Red, Solid).} Ten nodes exhibit degree $\ge 10$, indicating dense connectivity within the graph. These nodes tend to occupy central positions, as they serve as anchors connecting multiple conversation threads. The Reflective Agent identifies such high-degree nodes and applies Hebbian Distillation to consolidate their associated episodic clusters into stable semantic entries. For instance, the node with degree 17 in Figure \ref{fig:reflective_dual} links several temporally dispersed discussions, making it a natural candidate for knowledge extraction.

\textbf{Isolated Nodes (Gray, Dashed).} Four nodes exhibit degree $< 4$ and show no recent access activity. Their peripheral positions and weak integration suggest limited relevance to the ongoing narrative. The Adaptive Forgetting mechanism flags these nodes for removal, thereby controlling graph growth and reducing retrieval noise over extended conversations. The remaining nine nodes (blue) have moderate connectivity and are retained in the episodic store.

This visualization confirms that Hebbian learning enables automated lifecycle management without manual annotation (see Appendix \ref{app:heatmap} for the edge-weight heatmap).

\subsection{Case Study: Trace Analysis of Associative Recall}

We analyze the retrieval process for a multi-hop query: ``\textit{Where did you first meet the person who influenced your career choice?}'' (see Figure \ref{fig:case_study}).

\textbf{Historical Context.} The entities ``Dr. Sarah'' (Person) and ``Adoption Support Conference'' (Location) appeared together in Session 1, and were subsequently co-activated in Sessions 39 and 61. Through Hebbian learning, these repeated co-occurrences accumulated a strong associative weight of $w_{89,15} \approx 0.52$ between the career advice memory (Turn 89) and the meeting event (Turn 15). This accumulated weight reflects the frequency and recency of co-activation across the conversation history.

\textbf{Experimental Trace.} The baseline method identifies Turn 89 (``Dr. Sarah encouraged...'') as the top candidate due to high semantic similarity (0.82) but fails to retrieve Turn 15 (``Met Dr. Sarah at...'') because its low similarity score of 0.35 falls below the retrieval threshold. This results in a ``semantic trap'' where the model knows the person but not the location. 

In contrast, HeLa-Mem utilizes the Hebbian path. Spreading activation from the retrieved Turn 89 propagates through the learned edge ($0.52$) to Turn 15. The final retrieval score for Turn 15 is dynamically updated:
\begin{equation}
S_{total} = \underbrace{0.35}_{\text{Semantic}} + \underbrace{0.36}_{\text{Hebbian}} \approx \mathbf{0.71}
\end{equation}
This score boost, derived strictly from historical association, promotes Turn 15 into the active context. By retrieving both the cue (Person) and the target (Location), HeLa-Mem correctly synthesizes the answer: ``\textit{At the adoption support conference.}'' This demonstrates how Hebbian associations complement semantic retrieval for complex reasoning.

\section{Conclusion}

We introduce \textbf{HeLa-Mem}, a bio-inspired memory architecture that models conversation history as a dynamic graph driven by Hebbian learning principles. Unlike static context windows, HeLa-Mem mimics the brain's plasticity, where ``neurons that fire together, wire together,'' enabling the spontaneous emergence of associative pathways for retrieval. Building on this foundation, our Reflective Agent distills transient episodes into structured semantic knowledge, while Adaptive Forgetting ensures long-term scalability. Experiments on the LoCoMo benchmark demonstrate that this synergy between associative retention and semantic consolidation yields superior performance across diverse question types and robustness across diverse LLM backbones. These findings suggest that incorporating neuro-symbolic dynamics offers a promising direction for evolving static LLMs into lifelong learning agents.

\clearpage

\section{Limitations}
While HeLa-Mem effectively models long-term memory consolidation, it faces a ``cold start'' challenge: Hebbian weights require sufficient interaction history to accumulate, meaning the benefits of associative retrieval are less pronounced in early conversation stages. Future work may explore initializing Hebbian edges using semantic similarity as a prior, allowing the graph structure to bootstrap before sufficient co-occurrences accumulate. Additionally, the quality of both Semantic Memory and Hub Detection relies on the capabilities of the underlying LLM; hallucinations or reasoning errors during the distillation process could propagate into the long-term storage, potentially affecting future retrieval accuracy.

\section{Ethical Considerations}
We use the publicly available LoCoMo benchmark and do not collect any private user data. The proposed memory architecture is intended to enhance the consistency of LLM agents. However, we acknowledge that long-term memory systems could potentially reinforce biases present in the underlying LLM if not carefully monitored. The distilled semantic memories should be treated with the same caution as standard LLM generations regarding accuracy and bias.

\section*{Acknowledgments}
This work was partially supported by the Guangdong Provincial Natural Science Foundation General Program (Grant No. 2026A1515012118).

\bibliography{main}

\clearpage

\appendix

\lstdefinelanguage{json_blue}{
    basicstyle=\small\ttfamily,
    showstringspaces=false,
    breaklines=true,
    backgroundcolor=\color{mycolor_lightblue},
    commentstyle=\color{green}\ttfamily,
    keywordstyle=\color{blue}\ttfamily,
    stringstyle=\color{blue}\ttfamily,      
    keywordstyle={[2]\color{blue}\ttfamily},
    morekeywords={[2] 
        \{\%, \%\}, \{\{, \}\},
    },
    literate=
        {\{\%}{{\textcolor{blue}{\{\%}}}{1}
        {\%\}}{{\textcolor{blue}{\%\}}}}{1}
        {\{\{}{{\textcolor{blue}{\{\{~}}}{1}
        {\}\}}{{\textcolor{blue}{~\}\}}}}{1},
    sensitive=true,
    frame=single, 
    framesep=1em, 
    xleftmargin=1em, 
    xrightmargin=1em, 
    morekeywords={}, 
}

\lstdefinelanguage{json_green}{
    basicstyle=\small\ttfamily,
    showstringspaces=false,
    breaklines=true,
    backgroundcolor=\color{mycolor_lightgreen},
    commentstyle=\color{green}\ttfamily,
    keywordstyle=\color{blue}\ttfamily,
    stringstyle=\color{blue}\ttfamily,   
    keywordstyle={[2]\color{blue}\ttfamily},
    morekeywords={[2] 
        \{\%, \%\}, \{\{, \}\},
    },
    literate=
        {\{\%}{{\textcolor{blue}{\{\%}}}{1}
        {\%\}}{{\textcolor{blue}{\%\}}}}{1}
        {\{\{}{{\textcolor{blue}{\{\{~}}}{1}
        {\}\}}{{\textcolor{blue}{~\}\}}}}{1},
    sensitive=true,
    frame=single, 
    framesep=1em, 
    xleftmargin=1em, 
    xrightmargin=1em, 
    morekeywords={},   
}

\lstdefinelanguage{json_red}{
    basicstyle=\small\ttfamily,
    showstringspaces=false,
    breaklines=true,
    backgroundcolor=\color{mycolor_lightred},
    commentstyle=\color{green}\ttfamily,
    keywordstyle=\color{blue}\ttfamily,
    stringstyle=\color{blue}\ttfamily,    
    keywordstyle={[2]\color{blue}\ttfamily},
    morekeywords={[2] 
        \{\%, \%\}, \{\{, \}\},
    },
    literate=
        {\{\%}{{\textcolor{blue}{\{\%}}}{1}
        {\%\}}{{\textcolor{blue}{\%\}}}}{1}
        {\{\{}{{\textcolor{blue}{\{\{~}}}{1}
        {\}\}}{{\textcolor{blue}{~\}\}}}}{1},
    sensitive=true,
    frame=single, 
    framesep=1em, 
    xleftmargin=1em, 
    xrightmargin=1em, 
    morekeywords={},  
}

\lstdefinelanguage{json_pink}{
    basicstyle=\small\ttfamily,
    showstringspaces=false,
    breaklines=true,
    backgroundcolor=\color{mycolor_lightpink},
    commentstyle=\color{green}\ttfamily,
    keywordstyle=\color{blue}\ttfamily,
    stringstyle=\color{blue}\ttfamily,    
    keywordstyle={[2]\color{blue}\ttfamily},
    morekeywords={[2] 
        \{\%, \%\}, \{\{, \}\},
    },
    literate=
        {\{\%}{{\textcolor{blue}{\{\%}}}{1}
        {\%\}}{{\textcolor{blue}{\%\}}}}{1}
        {\{\{}{{\textcolor{blue}{\{\{~}}}{1}
        {\}\}}{{\textcolor{blue}{~\}\}}}}{1},
    sensitive=true,
    frame=single, 
    framesep=1em, 
    xleftmargin=1em, 
    xrightmargin=1em, 
    morekeywords={},  
}

\onecolumn

\section{LLM Prompts}
\label{app:prompts}

This appendix provides the core LLM prompts used in the HeLa-Mem system.

\subsection{Hebbian Distillation}

The Reflective Agent uses the following prompt to extract structured knowledge from memory clusters identified as hubs.

\begin{lstlisting}[caption=Semantic Memory Extraction Prompt, language=json_blue]
System: You are a knowledge extraction engine analyzing conversation memories.
Extract ONLY factual information with direct evidence.
Output concise, structured entries.

User: Analyze the following memory cluster and extract:

1. USER CHARACTERISTICS:
   - Observable traits (with evidence)
   - Content preferences (with evidence)
   - Interaction patterns

2. FACTUAL INFORMATION:
   - Events with dates and locations
   - Stated preferences
   - Mentioned relationships

Format: Concise bullet points with supporting evidence.

Memory Cluster: {conversation}
\end{lstlisting}

\subsection{Response Generation}

The system uses the following prompt to generate responses using retrieved episodic and semantic memories.

\begin{lstlisting}[caption=Response Generation Prompt, language=json_pink]
System: You are an AI assistant with access to conversation history.
Answer questions concisely using the provided context.
For dates, use format "15 July 2023".

User: 
<EPISODIC MEMORIES>
{episodic_context}

<SEMANTIC KNOWLEDGE>
{semantic_knowledge}

<USER CHARACTERISTICS>
{user_model}

Question: {query}

Provide an extremely concise answer using concrete entities.
Output only the answer content, without labels.
\end{lstlisting}

\twocolumn

\section{Hebbian Weight Visualization}
\label{app:heatmap}

\begin{figure}[h]
    \centering
    \includegraphics[width=0.9\linewidth]{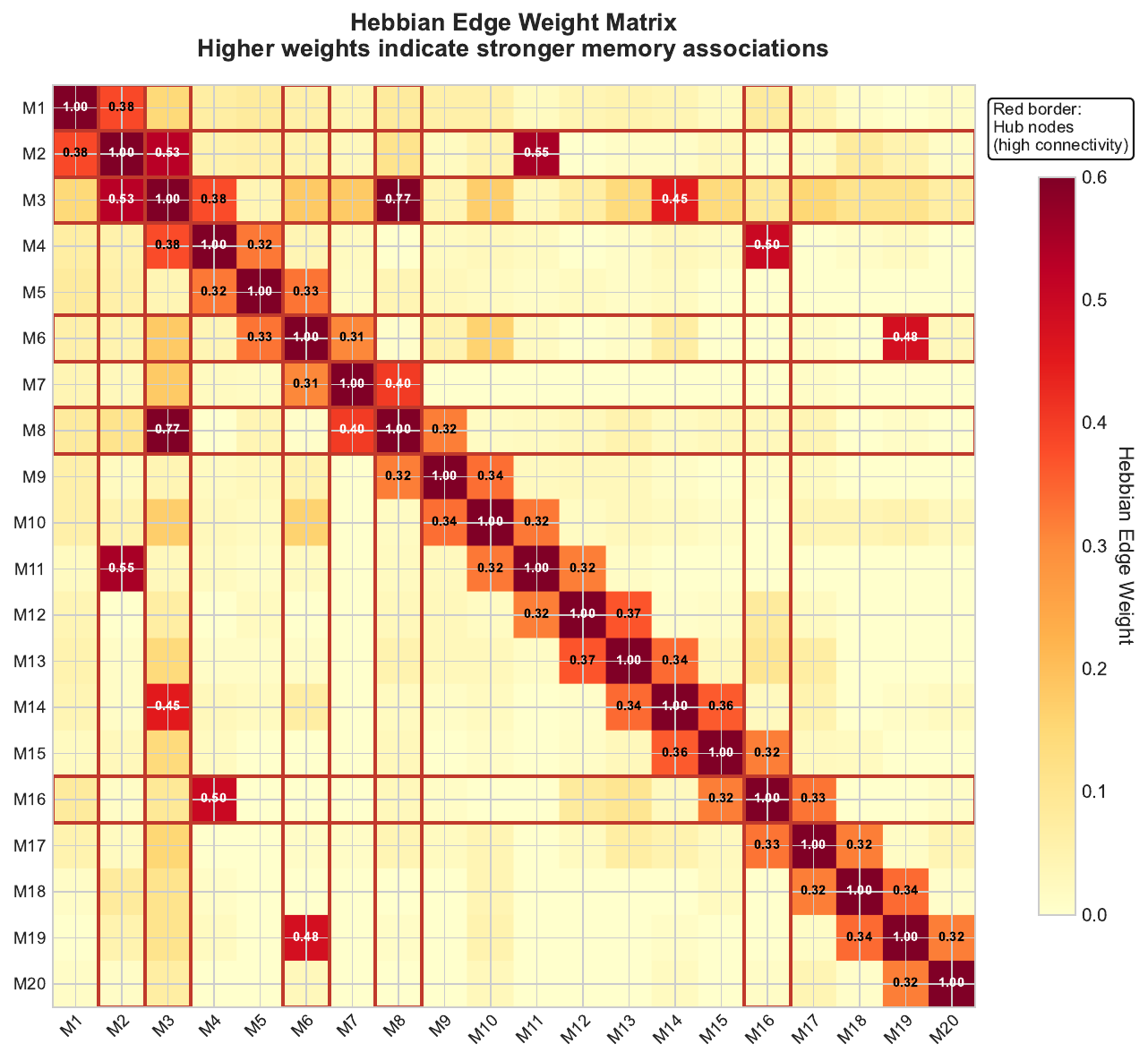}
    \caption{Hebbian edge weight matrix for the first 20 memory nodes. Stronger weights (darker colors) indicate more frequent co-activation. Nodes with red borders have high total connectivity across multiple memories.}
    \label{fig:heatmap}
\end{figure}

Figure \ref{fig:heatmap} shows the Hebbian edge weight matrix for the first 20 memory nodes. Stronger weights (darker colors) indicate associations formed through co-activation. The matrix exhibits both local associations near the diagonal and cross-topic connections between distant nodes, demonstrating that Hebbian learning captures semantic relationships beyond temporal adjacency. Nodes with red borders have high total connectivity, making them candidates for Hebbian Distillation.

\section{Dataset Statistics}
\label{app:dataset}

We utilize the LoCoMo benchmark, focusing on the long-context conversation split. Table \ref{tab:dataset_stats} provides the detailed statistics of the 10 conversations used in our experiments, while Table \ref{tab:question_types} details the distribution across different question categories.

\begin{table}[!ht]
    \centering
    \begin{minipage}[t]{0.48\textwidth}
        \centering
        \caption{LoCoMo dataset overview.}
        \label{tab:dataset_stats}
        \small
        \begin{tabular}{>{\raggedright\arraybackslash}p{4.5cm} | >{\centering\arraybackslash}p{1.5cm}}
        \toprule
        \textbf{Metric} & \textbf{Value} \\
        \midrule
        Number of Conversations & $10$ \\
        Avg. Turns per Conversation & $\sim 300$ \\
        Avg. Tokens per Conversation & $\sim 9,000$ \\
        Total Question-Answer Pairs & $1,986$ \\
        \bottomrule
        \end{tabular}
    \end{minipage}
    \hfill
    \vspace{1em}
    \begin{minipage}[t]{0.48\textwidth}
        \centering
        \caption{Distribution of question types in the evaluation set.}
        \label{tab:question_types}
        \small
        \begin{tabular}{>{\raggedright\arraybackslash}p{2.8cm} | 
            >{\centering\arraybackslash}p{1.5cm} | 
            >{\centering\arraybackslash}p{1.8cm}}
        \toprule
        \textbf{Category} & \textbf{Count} & \textbf{Percentage} \\
        \midrule
        Single-hop & $841$ & $42.3\%$ \\
        Multi-hop & $282$ & $14.2\%$ \\
        Temporal & $321$ & $16.2\%$ \\
        Open-domain & $96$ & $4.8\%$ \\
        Adversarial & $446$ & $22.5\%$ \\
        \midrule
        \textbf{Total} & $\mathbf{1,986}$ & $\mathbf{100.0\%}$ \\
        \bottomrule
        \end{tabular}
    \end{minipage}
\end{table}

\section{Additional Benchmark: LongMemEval-S}
\label{app:longmemeval}

We additionally evaluate HeLa-Mem on LongMemEval-S, a 500-item long-term conversational memory benchmark. We use GPT-4o-mini as both the backbone model and the LLM judge. For retrieval, HeLa-Mem uses top-15 episodic memories and top-5 semantic memories (20 total). The best setting on this benchmark uses learning rate $\eta=0.02$, decay rate $\lambda=0.995$, spreading activation strength $\beta=0.1$, spreading threshold $\theta=0.4$, keyword weight $0.7$, and max flipped items $m=3$. In Table~\ref{tab:longmemeval_category}, \textbf{Single} denotes the merged single-hop group combining Single-User, Single-Asst, and Single-Pref. Baseline numbers are reported from \citet{fang2025lightmem} under the same total retrieval budget.

\begin{table}[!ht]
    \centering
    \caption{Overall accuracy on LongMemEval-S.}
    \label{tab:longmemeval_overall}
    \small
    \begin{tabular}{>{\raggedright\arraybackslash}p{4.2cm} | >{\centering\arraybackslash}p{1.7cm}}
    \toprule
    \textbf{Method} & \textbf{ACC (\%)} \\
    \midrule
    LangMem & 37.20 \\
    MemoryOS & 44.80 \\
    Mem0 & 53.61 \\
    FullText & 56.80 \\
    NaiveRAG & 61.00 \\
    A-MEM & 62.60 \\
    \textbf{HeLa-Mem} & \textbf{65.40} \\
    \bottomrule
    \end{tabular}
\end{table}

\begin{table}[!htbp]
    \centering
    \caption{Category-wise accuracy on LongMemEval-S.}
    \label{tab:longmemeval_category}
    \scriptsize
    \setlength{\tabcolsep}{4pt}
    \resizebox{\columnwidth}{!}{
    \begin{tabular}{l|c|c|c|c}
    \toprule
    \textbf{Method} & \textbf{Temporal} & \makecell{\textbf{Multi-}\\\textbf{Session}} & \makecell{\textbf{Knowledge-}\\\textbf{Update}} & \textbf{Single} \\
    \midrule
    LangMem & 15.79 & 20.30 & 66.67 & 55.13 \\
    MemoryOS & 32.33 & 31.06 & 48.72 & 64.74 \\
    Mem0 & 40.15 & 46.21 & 70.12 & 62.82 \\
    FullText & 31.58 & 45.45 & 76.92 & 78.21 \\
    NaiveRAG & 39.85 & 48.48 & 67.95 & \textbf{85.90} \\
    A-MEM & 47.36 & 48.87 & 64.11 & 84.62 \\
    \textbf{HeLa-Mem} & \textbf{50.38} & \textbf{57.14} & \textbf{78.21} & 78.85 \\
    \bottomrule
    \end{tabular}
    }
\end{table}

\FloatBarrier

\noindent
The category-wise values are not directly averaged to obtain the overall ACC because the category sizes are unequal.

\noindent
HeLa-Mem achieves the best overall accuracy and the best performance on the three reasoning-intensive categories: Temporal, Multi-Session, and Knowledge-Update.

\section{LLM Usage Statement}

We use publicly available large language model tools as writing assistants to check grammar and polish a small number of sentences. All technical content, claims, and contributions are conceived, written, and verified by the authors. For schematic figures, several icons or visual elements are refined with the assistance of LLM-based design tools, while the figure layout, semantics, and interpretation are fully determined by the authors. Since this paper involves LLM-related research, all model usage that affects experiments, analysis, or results is explicitly documented in Section Experiments. No other parts of the manuscript are generated or substantively rewritten by an LLM.

\end{document}